%% file: 00_main.tex
\documentclass{article}
\usepackage[utf8]{inputenc}
\usepackage{amsfonts}
\usepackage[mathscr]{euscript}
\usepackage{graphicx}
\usepackage[numbers]{natbib} 
\usepackage{mathtools} 
\usepackage{bbm}
\usepackage{boxedminipage}
\usepackage{alltt}
\usepackage{bm}
\usepackage{geometry}
\usepackage{hyperref}
\usepackage{authblk}
\usepackage{lineno}         
\usepackage{lipsum}
\usepackage{amssymb}

\DeclareMathOperator{\defeq}{\stackrel{\text{def}}{=}}

\title{Double Descent Demystified: Identifying, Interpreting \& Ablating the Sources of a Deep Learning Puzzle}
\author[1]{Rylan Schaeffer}
\author[2]{Mikail Khona}
\author[1]{Zachary Robertson}
\author[3]{Akhilan Boopathy}
\author[4]{Kateryna Pistunova}
\author[5]{Jason W. Rocks}
\author[6]{Ila Rani Fiete}
\author[1]{Oluwasanmi Koyejo}

\affil[1]{Computer Science, Stanford University}
\affil[2]{Physics, Massachusetts Institute of Technology}
\affil[3]{EECS, Massachusetts Institute of Technology}
\affil[4]{Physics, Stanford University}
\affil[5]{Physics, Boston University}
\affil[6]{Brain \& Cognitive Sciences, Massachusetts Institute of Technology}

\date{March 2023}

\begin{document}

\maketitle

\begin{abstract}
    Double descent is a surprising phenomenon in machine learning, in which as the number of model parameters grows relative to the number of data, test error drops as models grow ever larger into the highly overparameterized (data undersampled) regime. This drop in test error flies against classical learning theory on overfitting and has arguably underpinned the success of large models in machine learning. This non-monotonic behavior of test loss depends on the number of data, the dimensionality of the data and the number of model parameters.
    Here, we briefly describe double descent, then provide an explanation of why double descent occurs in an informal and approachable manner, requiring only familiarity with linear algebra and introductory probability.
    We provide visual intuition using polynomial regression, then mathematically analyze double descent with ordinary linear regression and identify three interpretable factors that, when simultaneously all present, together create double descent.
    We demonstrate that double descent occurs on real data when using ordinary linear regression, then demonstrate that double descent does not occur when any of the three factors are ablated.
    We use this understanding to shed light on recent observations in nonlinear models concerning superposition and double descent.
    Code is \href{https://github.com/RylanSchaeffer/Stanford-AI-Alignment-Double-Descent-Tutorial}{publicly available}.
\end{abstract}


\input{01_introduction}
\input{02_polynomial_regression}

\input{03_linear_regression}
\input{04_nonlinear}

\input{05_discussion}

\clearpage

\bibliographystyle{plain}
\bibliography{references}

\clearpage

\appendix

\input{Implicit_Bias_GD}

\end{document}

%% file: 01_introduction.tex
\section{What is double descent?}

\begin{figure}
    \centering
    \includegraphics[width=0.6\textwidth]{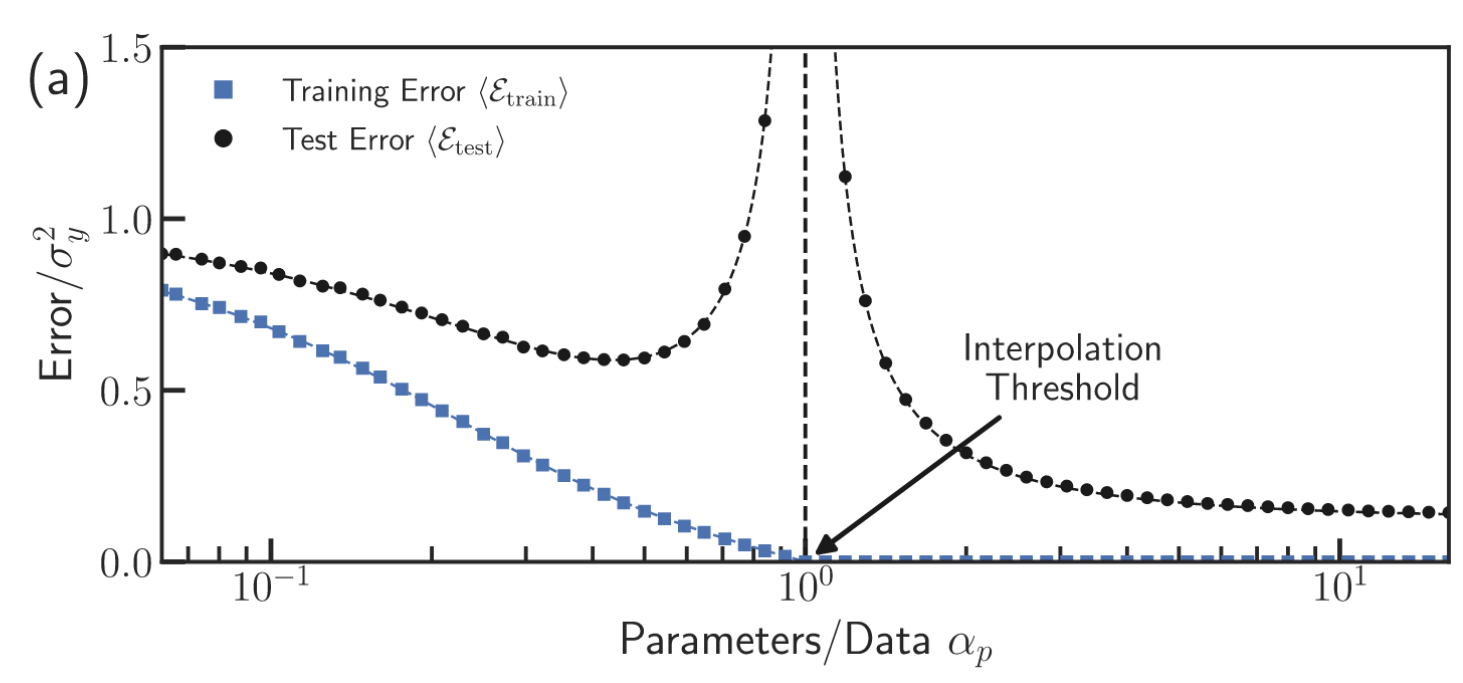}
    \caption{Double descent: test error falls, rises, then falls as a ratio of parameters to data. Fig. 3A from \cite{rocks2022memorizing}.}
    \label{fig:rocks2022memorizing}
\end{figure}

Double descent is a phenomenon in machine learning describing the key observation that many classes of models can, under relatively broad conditions, exhibit seemingly perplexing changes in test loss as a function of three parameters: the number of data, the dimensionality of the data and the number of parameters in the model.
For instance, as the number of model parameters increases, the test loss can fall, then rise, then fall again (Fig. \ref{fig:rocks2022memorizing}).
What causes such behavior? \newline

Double descent has a rich history, both empirically and analytically; for a non-exhaustive list, see \citep{opper1995statistical, advani2020high, spigler2018jamming, nakkiran2021deep, poggio2019double, advani2020high, adlam2020understanding, mei2022generalization, hastie2022surprises}.
The term ``double descent" was coined by \cite{belkin2019reconciling}, and some of our favorite papers on the topic include \cite{rocks2021geometry, rocks2022bias, rocks2022memorizing}.
One important note is that not every dataset and model pair exhibits \textit{two} descents; under different settings, there can be one, three or more descents \cite{nakkiran2020optimal,chen2021multiple}.

Our goal in this tutorial is to explain why double descent occurs in an approachable manner, without resorting to advanced tools oftentimes employed to analyze double descent such as random matrix theory or statistical physics.
To accomplish this, we provide (1) visual intuition via polynomial regression, (2) mathematical analysis using ordinary linear regression, (3) empirical evidence from ordinary linear regression on real (tabular) data and (4) novel insights for nonlinear neural networks.
To the best of our knowledge, we are the first to take this approach.
Although we focus on regression tasks, our insights hold more generally.

\section{Notation and Terminology}

Consider a supervised dataset of $N$ training data for regression:
\begin{equation*}
    \mathcal{D} \, \defeq \, \Big\{ (\vec{x}_n, y_n) \Big\}_{n=1}^N
\end{equation*}

with covariates $\vec{x}_n \in \mathbb{R}^D$ and targets $y_n \in \mathbb{R}$.
We'll sometimes use matrix-vector notation to refer to our training data, treating the features $\vec{x}_n$ as row vectors:
\begin{equation*}
    X \, \defeq \, \begin{bmatrix} - \vec{x}_1 - \\ 
    \vdots\\
    - \vec{x}_N - \end{bmatrix} \in \mathbb{R}^{N \times D}
    \quad \quad \quad \quad 
    Y \, \defeq \, \begin{bmatrix} y_1\\  \vdots \\ y_N \end{bmatrix} \in \mathbb{R}^{N \times 1}   
\end{equation*}

In general, our goal is to use our training dataset $\mathcal{D}$ find a function $f: \mathcal{X} \rightarrow \mathcal{Y}$ that makes:
\begin{equation*}
    f(x) \approx y   
\end{equation*}

In the setting of ordinary linear regression, we assume that $f$ is a linear function i.e. $f(\vec{x}) = \vec{x} \cdot \vec{\beta}$, meaning our goal is to find (estimate) linear parameters $\hat{\vec{\beta}} \in \mathbb{R}^{D}$ that make:
\begin{equation*}
    \vec{x} \cdot \vec{\beta} \approx y
\end{equation*}

Of course, our real goal is to hopefully find a function that generalizes well to new data. As a matter of terminology, there are typically three key parameters:

\begin{enumerate}
    \item The number of model parameters $P$
    \item The number of training data $N$
    \item The dimensionality of the data $D$
\end{enumerate}

We say that a model is \textit{overparameterized} (a.k.a. underconstrained) if $N < P$ and \textit{underparameterized} (a.k.a. overconstrained) if $N > P$.
The \textit{interpolation threshold} refers to where $N=P$, because when $P\geq N$, the model can perfectly interpolate the training points.

%% file: 02_polynomial_regression.tex
\section{Visual Intuition from Polynomial Regression}

\begin{figure}[h]
    \centering
    \includegraphics[width=0.99\textwidth]{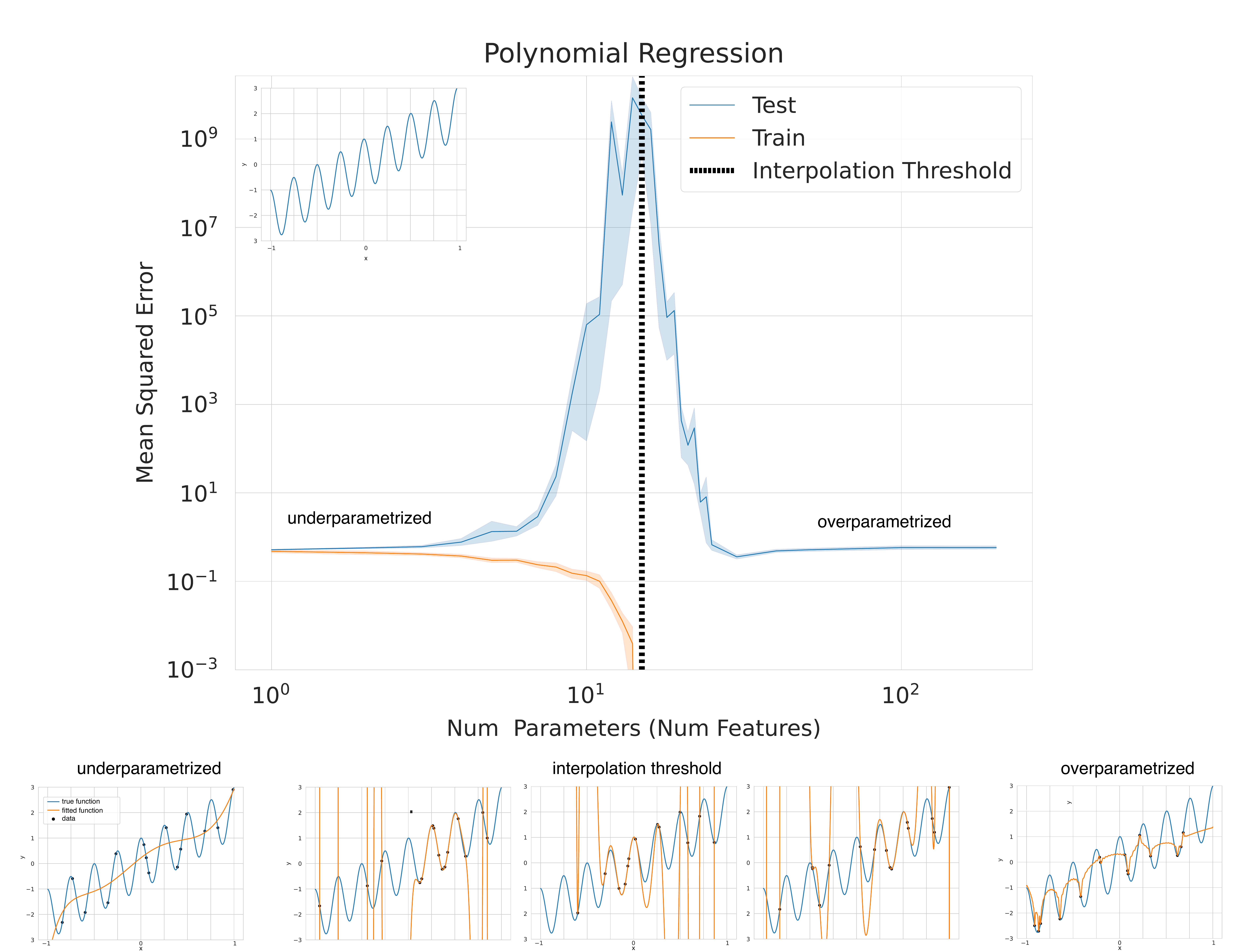}
    \caption{\textbf{Intuition for double descent from polynomial regression.} Top: Polynomial regression displays double descent.
    Bottom: When \textit{underparameterized}, the model is unable to capture finer-grained features in the training data, meaning bias is large but variance is small. As the interpolation threshold is approached, the training data can be fit exactly, meaning bias is small; however, the particular realization of the training data significantly affects the learnt function, meaning variance is large. When \textit{overparameterized}, the model can exactly fit the training data, meaning bias is again small, but the model is also regularized towards a small-norm solution, making variance small.}
    \label{fig:polynomial_regression}
\end{figure}

To offer visual intuition for the cause of double descent, we turn to polynomial regression.
Concretely, suppose we wish to predict $y \in \mathbb{R}$ from $x \in [-1, 1]$, where the true (unknown) relationship is:
\begin{equation*}
    y(x) = 2 x + \cos(25 x)
\end{equation*}

In polynomial regression, we take the approach of mapping each datum $x$ to a $P$-dimensional space (corresponding to the $P$ parameters) by using the following ``feature" map $\vec{\phi}_P: \mathbb{R}^1 \rightarrow \mathbb{R}^P$:
\begin{equation*}
    \vec{\phi}_P: x \rightarrow \begin{bmatrix}
        \phi_1(x)\\
        \phi_2(x)\\
        \vdots\\
        \phi_P(x)
    \end{bmatrix} \in \mathbb{R}^P
\end{equation*}

where $\phi_i$ denotes some polynomial\footnote{In our simulations, we choose $\phi_i$ to be the $i$-th Legendre polynomial}. We'll then fit a linear model using parameters $\vec{\beta}_P \in \mathbb{R}^P$:
\begin{equation*}
    y \approx \vec{\phi}_P(x) \cdot \vec{\beta}_P    
\end{equation*}

We show the results of sweeping the number of parameters $P$ (= the number of polynomials = the number of features) from 1 to 200 (Fig. \ref{fig:polynomial_regression}). 
When $P=1$, we can fit a line to our data, which is insufficiently expressive to capture the true relationship between $x$ and $y$; thus, the model has bias, but low variance.
As the number of parameters $P$ increases towards the number of training data $N$, the model can more accurately fit the training data but at the cost of inducing  ``wiggles" that depend on the particular realization of the training data and that incur a high test mean squared error; the bias decreases but the variance diverges.
As the number of parameters $P$ increases beyond the number of training data $N$, the regression remains sufficiently expressive to exactly fit the training data, meaning it has low bias, but the model fitting process prefers solutions with smaller norm that ``wiggle`` less, meaning it has low variance as well. Next, we mathematically analyze under what conditions this double-descent phenomenon occurs.

%% file: 03_linear_regression.tex
\section{Mathematical Intuition from Ordinary Linear Regression}

To offer an intuitive yet quantitative understanding of double descent, we turn to ordinary linear regression. Recall that in linear regression, the number of fit parameters $P$ must equal the dimension $D$ of the covariates; consequently, rather than thinking about changing the number of parameters $P$, we'll instead think about changing the number of data $N$. \textit{Because double descent is fundamentally about the ratio of number of parameters $P$ to number of data $N$}, varying the number of data is as valid an approach as varying the number of parameters is. To understand where and why double descent occurs in linear regression, we'll study how linear regression behaves in the two parameterization regimes. \newline

\begin{figure}
    \centering
    \includegraphics[width=0.48\textwidth]{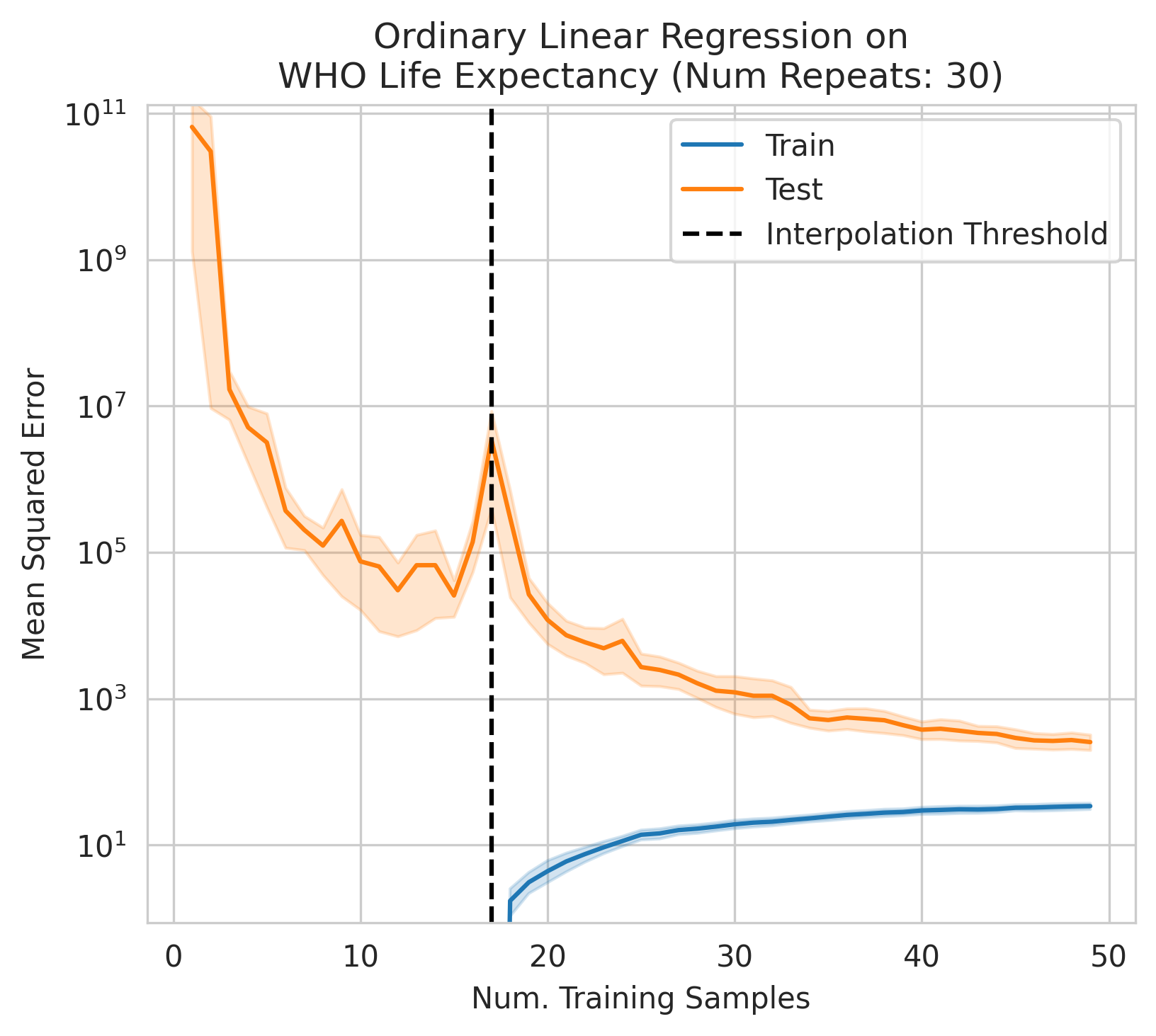}%
    \includegraphics[width=0.49\textwidth]{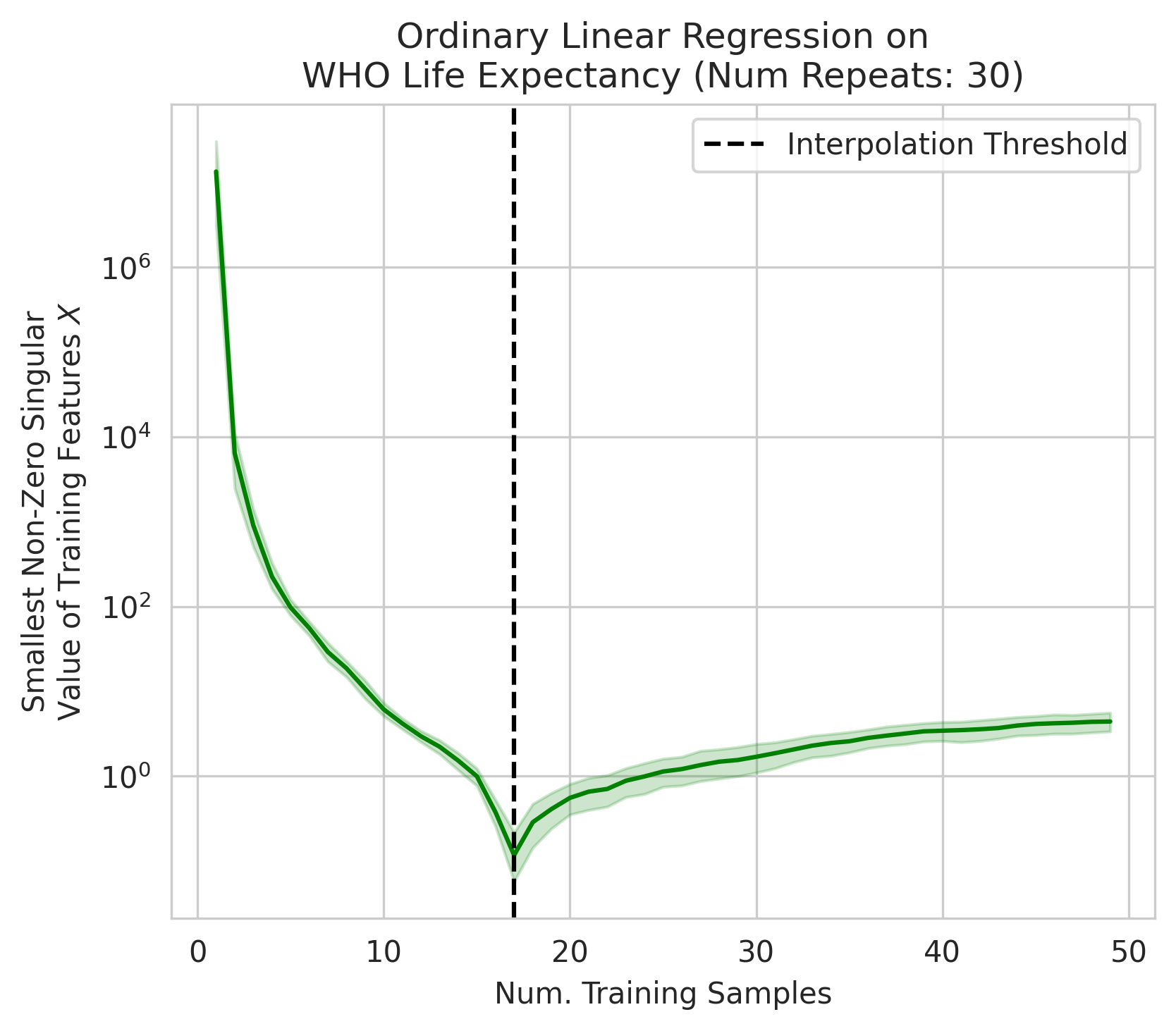}
    \caption{Left: Double descent occurs in ordinary linear regression on the World Health Organization's Life Expectancy dataset. Right: Double descent occurs when three quantities simultaneously grow extreme. One of the three is small-but-nonzero singular values in the training features $X$. The smallest non-zero singular value of $X$ (probabilistically) obtains its lowest value when the number of parameters $P$ equals the number of data $N$ (the \textit{interpolation threshold}), approaching from either the \textit{overparameterized regime} ($P > N$; left) or the \textit{underparameterized regime} ($P < N$; right).}
    \label{fig:WHO_life_expectancy}
\end{figure}

If the regression is \textit{underparameterized}, we estimate the linear relationship between the covariates $\vec{x}_n$ and the target $y_n$ by solving the classical least-squares minimization problem:
\begin{equation*}
    \hat{\vec{\beta}}_{under} \, \defeq \,  \arg \min_{\vec{\beta}} \frac{1}{N} \sum_n ||\vec{x}_n \cdot \vec{\beta} - y_n||_2^2 \, = \, \arg \min_{\vec{\beta}} ||X \vec{\beta} - Y ||_F^2
\end{equation*}

The solution to this underparameterized optimization problem is the well-known ordinary least squares estimator that uses the second moment matrix $X^T X$:
\begin{equation*}
    \hat{\vec{\beta}}_{under} = (X^T X)^{-1} X^T Y
\end{equation*}

If the model is \textit{overparameterized}, the above optimization problem is ill-posed since there are infinitely many solutions; this is because we have fewer constraints than parameters. Consequently, we need to choose a different (constrained) optimization problem:
\begin{equation*}
    \hat{\vec{\beta}}_{over} \, \defeq \, \arg \min_{\vec{\beta}} ||\vec{\beta}||_2^2 \quad \quad \text{s.t.} \quad \quad \forall \, n \in \{1, ..., N\} \quad \vec{x}_n \cdot \vec{\beta} = y_n
\end{equation*}

This constrained optimization problem asks for the smallest parameters $\vec{\beta}$ that guarantee $\vec{x}_n \cdot \vec{\beta} = y_n$ for all training data. One reason why we choose this optimization problem is that it is the optimization problem that gradient descent implicitly minimizes (App. \ref{app:why_sgd_regularizes}). The solution to this optimization problem is less well-known and instead uses the so-called \href{https://en.wikipedia.org/wiki/Gram_matrix}{Gram matrix} $X X^T \in \mathbb{R}^{N \times N}$:
\begin{equation*}
    \hat{\vec{\beta}}_{over} = X^T (X X^T)^{-1} Y
\end{equation*}

One way to see why the Gram matrix appears is via constrained optimization. Define the Lagrangian with Lagrange multipliers $\vec{\lambda} \in \mathbb{R}^N$:
\begin{equation*}
    \mathcal{L}(\vec{\beta}, \vec{\lambda}) \, \defeq \, ||\vec{\beta}||_2^2 + \vec{\lambda}^T (Y - X \vec{\beta})
\end{equation*}

Differentiating with respect to both the parameters and the Lagrange multipliers yields:
\begin{align*}
    \nabla_{\vec{\beta}}\,  \mathcal{L}(\vec{\beta}, \vec{\lambda}) = \vec{0} = 2\hat{\vec{\beta}} - X^T \vec{\lambda} &\Rightarrow \hat{\vec{\beta}}_{over} = \frac{1}{2} X^T \vec{\lambda}\\
    \nabla_{\vec{\lambda}} \,\mathcal{L}(\beta, \lambda) = \vec{0} = Y - X \hat{\vec{\beta}}_{over} &\Rightarrow Y = \frac{1}{2} X X^T \vec{\lambda}\\
    &\Rightarrow \vec{\lambda} = 2 (X X^T)^{-1} Y\\
    &\Rightarrow \hat{\vec{\beta}}_{over} = X^T (X X^T)^{-1} Y
\end{align*}

Here, we are able to invert the Gram matrix because it is full rank in the overparametrized regime.
After fitting its parameters, the model will make the following predictions for given test point $\vec{x}_{test}$:
\begin{align*}
    \hat{y}_{test, under} &= \vec{x}_{test} \cdot \hat{\Vec{\beta}}_{under} = \vec{x}_{test} \cdot (X^T X)^{-1} X^T Y\\
    \hat{y}_{test, over} &= \vec{x}_{test} \cdot \hat{\Vec{\beta}}_{over} 
    = \vec{x}_{test} \cdot X^T (X X^T)^{-1} Y
\end{align*}

\textit{Hidden in the above equations is an interaction between three quantities that can, when all grow extreme, create double descent.} To reveal the three quantities, we'll rewrite the regression targets by introducing a slightly more detailed notation. Unknown to us, there are some ideal linear parameters $\vec{\beta}^* \in \mathbb{R}^P = \mathbb{R}^D$ that truly minimize the test mean squared error. We can write any regression target as the inner product of the data $\vec{x}_n$ and the ideal parameters $\beta^*$, plus an additional error term $e_n$ that is an ``uncapturable" residual from the ``perspective" of the model class:
\begin{equation*}
    y_n = \vec{x}_n \cdot \vec{\beta}^* + e_n
\end{equation*}

In matrix-vector form, we will equivalently write:
\begin{equation*}
    Y = X \vec{\beta}^* + E
\end{equation*}

with $E \in \mathbb{R}^{N \times 1}$. To be clear, we are \textit{not} imposing assumptions on the model or data. Rather, we are introducing notation to express that there are (unknown) ideal linear parameters, and possibly residuals that even the ideal model might be unable to capture; these residuals could be random noise or could be fully deterministic patterns that this particular model class cannot capture. Using this new notation, we rewrite the model's predictions to show how the test datum's features $\vec{x}_{test}$, training data's features $X$ and training data's regression targets $Y$ interact. In the underparameterized regime:
\begin{align*}
    \hat{y}_{test,under} &= \Vec{x}_{test} \cdot (X^T X)^{-1} X^T Y\\
    &= \Vec{x}_{test} \cdot (X^T X)^{-1} X^T (X \beta^* + E)\\
    &= \Vec{x}_{test} \cdot (X^T X)^{-1} X^T X \beta^* + \vec{x}_{test} \cdot (X^T X)^{-1} X^T E\\
    &= \underbrace{\Vec{x}_{test} \cdot \beta^*}_{\defeq y_{test}^*} + \, \vec{x}_{test} \cdot (X^T X)^{-1} X^T E\\
    \hat{y}_{test,under} - y_{test}^* &= \vec{x}_{test} \cdot (X^T X)^{-1} X^T E
\end{align*}

This equation is important, but opaque. To extract the intuition, we will replace $X$ with its \href{https://en.wikipedia.org/wiki/Singular_value_decomposition}{singular value decomposition}\footnote{For those unfamiliar with the SVD, any real-valued $X$ can be decomposed into the product of three matrices $X = U \Sigma V^T$ where $U$ and $V$ are both orthonormal matrices and $\Sigma$ is diagonal; intuitively, any linear transformation can be viewed as composition of first a rotoreflection, then a scaling, then another rotoreflection. Because $\Sigma$ is diagonal and because $U, V$ are orthonormal matrices, we can equivalently write $X = U \Sigma V^T$ in vector-summation notation as a sum of rank-1 outer products $X = \sum_{r=1}^{rank(X)} \sigma_r u_r v_r^T$. Each term in the sum is referred to as a ``singular mode", akin to eigenmodes.} $X = U \Sigma V^T$ to reveal how different quantities interact. Let $R \, \defeq \, rank(X)$ and let $\sigma_1 > \sigma_2 > ... > \sigma_R > 0$ be $X$'s (non-zero) singular values. Recalling $E \in \mathbb{R}^{N \times 1}$, we can decompose the (underparameterized) prediction error $\hat{y}_{test, under} - y_{test}^*$ along the orthogonal singular modes:
\begin{align*}
    \hat{y}_{test, under} - y_{test}^* &= \Vec{x}_{test} \cdot V \Sigma^{+} U^T E = \sum_{r=1}^R  \frac{1}{\sigma_r} (\Vec{x}_{test} \cdot \vec{v}_r) (\vec{u}_r \cdot E)
\end{align*}

In the overparameterized regime, our calculations change slightly:
\begin{align*}
    \hat{y}_{test,over} &= \Vec{x}_{test} \cdot X^T (X X^T)^{-1}  Y\\
    &= \vec{x}_{test} \cdot X^T (X X^T)^{-1} (X \beta^* + E)\\
    &= \vec{x}_{test} \cdot X^T (X X^T)^{-1} X \beta^* + \vec{x}_{test} \cdot X^T (X X^T)^{-1} E\\
    \hat{y}_{test,over} - \underbrace{\vec{x}_{test} \cdot \beta^*}_{\defeq y_{test}^*} &= \vec{x}_{test} \cdot X^T (X X^T)^{-1} X \beta^*  - \vec{x}_{test} \cdot I_{D} \beta^* + \vec{x}_{test} \cdot (X^T X)^{-1} X^T E\\
    \hat{y}_{test,over} - y_{test}^* &= \vec{x}_{test} \cdot (X^T (X X^T)^{-1} X - I_D) \beta^*  + \vec{x}_{test} \cdot (X^T X)^{-1} X^T E
\end{align*}

If we again replace $X$ with its SVD $U S V^T$, we can again simplify $\vec{x}_{test} \cdot (X^T X)^{-1} X^T E$. This yields our final equations for the prediction errors.
\begin{align*}
\hat{y}_{test,over} - y_{test}^* &= \vec{x}_{test} \cdot (X^T (X X^T)^{-1} X - I_D) \beta^* \quad \quad \quad \quad + && \sum_{r=1}^R  \frac{1}{\sigma_r} (\Vec{x}_{test} \cdot \vec{v}_r) (\vec{u}_r \cdot E)\\
    \hat{y}_{test,under} - y_{test}^* &= &&\sum_{r=1}^R  \frac{1}{\sigma_r} (\Vec{x}_{test} \cdot \vec{v}_r) (\vec{u}_r \cdot E)
\end{align*}

What is the discrepancy between the underparameterized prediction error and the overparameterized prediction error, and from where does the discrepancy originate? The overparameterized prediction error $\hat{y}_{test,over} - y_{test}^*$ has the extra term $\vec{x}_{test} \cdot (X^T (X X^T)^{-1} X - I_D) \beta^*$. To understand where this term originates, recall that our goal is to understand how fluctuations in the features $\vec{x}$ correlate with fluctuations in the targets $y$. In the overparameterized regime, there are more parameters than there are data. Consequently, for $N$ data points in $D=P$ dimensions, the model can ``see" fluctuations in at most $N$ dimensions, but has no ``visibility" into fluctuations in the remaining $P-N$ dimensions. This causes information about the optimal linear relationship $\vec{\beta}^*$ to be lost, which in turn increases the overparameterized prediction error $\hat{y}_{test, over} - y_{test}^*$. Statisticians call this term $\vec{x}_{test} \cdot (X^T (X X^T)^{-1} X - I_D) \beta^*$ the ``bias". The other term (the ``variance") is what causes double descent:
\begin{equation}
    \sum_{r=1}^R  \frac{1}{\sigma_r} (\Vec{x}_{test} \cdot \vec{v}_r) (\vec{u}_r \cdot E)
    \label{eqn:variance}
\end{equation}

\textit{Eqn. \ref{eqn:variance} is critical}. It reveals that our test prediction error (and thus, our test squared error!) will depend on an interaction between 3 quantities:
\begin{enumerate}
    \item How much the \textit{training features} $X$ vary in each direction; more formally, the inverse (non-zero) singular values of the \textit{training features} $X$:
    $$\frac{1}{\sigma_r}$$
    
    \item How much, and in which directions, the \textit{test features} $\vec{x}_{test}$ vary relative to the \textit{training features} $X$; more formally: how $\vec{x}_{test}$ projects onto $X$'s right singular vectors $V$:
    $$\Vec{x}_{test} \cdot \Vec{v}_r$$
    
    \item How well the \textit{best possible model} can correlate the variance in the \textit{training features} $X$ with the \textit{training regression targets} $Y$; more formally: how the residuals $E$ of the best possible model in the model class (i.e. insurmountable ``errors" from the ``perspective" of the model class) project onto $X$'s left singular vectors $U$:
    $$\Vec{u}_r \cdot E$$
    
\end{enumerate}

Here, we use the terminology ``vary" and ``variance", suggesting a connection to \href{https://en.wikipedia.org/wiki/Covariance_matrix}{the statistical notion of variance}. There is indeed one\footnote{If we had centered the training features $X$ to form $\bar{X}$, with corresponding SVD $\bar{X} = \bar{U} \bar{\Sigma} \bar{V}^T$, then $\bar{X}^T \bar{X}$ would be the empirical covariance matrix, and its eigendecomposition would be:
\begin{equation*}
    \bar{X}^T \bar{X} = \bar{V} \bar{\Sigma}^2 \bar{V}^T   
\end{equation*}

Each right singular vector of $\bar{X}$ would be an orthogonal axis of variation, with the variance along each direction given by the squared singular values of $\bar{X}$. However, because we don't center the features, $X^T X$ is the empirical second moment matrix, not the empirical covariance matrix. Thus, when we refer to ``vary" and ``variance", we are slightly abusing terminology, but the intuition - how much the features are wiggling, and whether the wiggling is correlated with the regression targets - is the right way to understand the concepts. We could center our data features to refer to the variance, but we felt doing so might be pedagogically confusing since centering is not related to double descent.}, although we slightly abuse terminology! These three quantities, multiplied, determine how much the $r$-th singular mode contributes to the prediction error. \newline

Double descent occurs when these three quantities grow extreme: (i) the \textit{training features} contain small-but-nonzero variance in some singular direction(s), (ii) from the ``perspective" of the model class, residual errors in the \textit{training features and targets} have large projection along this singular mode, and (iii) the \textit{test features} vary significantly along this singular mode. When (i) and (ii) co-occur, this means the model's parameters along this mode are likely incorrect. Then, when (iii) is added to the mix by a test datum $\vec{x}_{test}$ with a large projection along this mode, the model is forced to extrapolate significantly beyond what it saw in the training data, in a direction where the training data had an error-prone relationship between its training predictions and the training regression targets, using parameters that are likely wrong. As a consequence, the test squared error explodes! \newline


\begin{figure}
    \centering
    \includegraphics[width=0.32\textwidth]{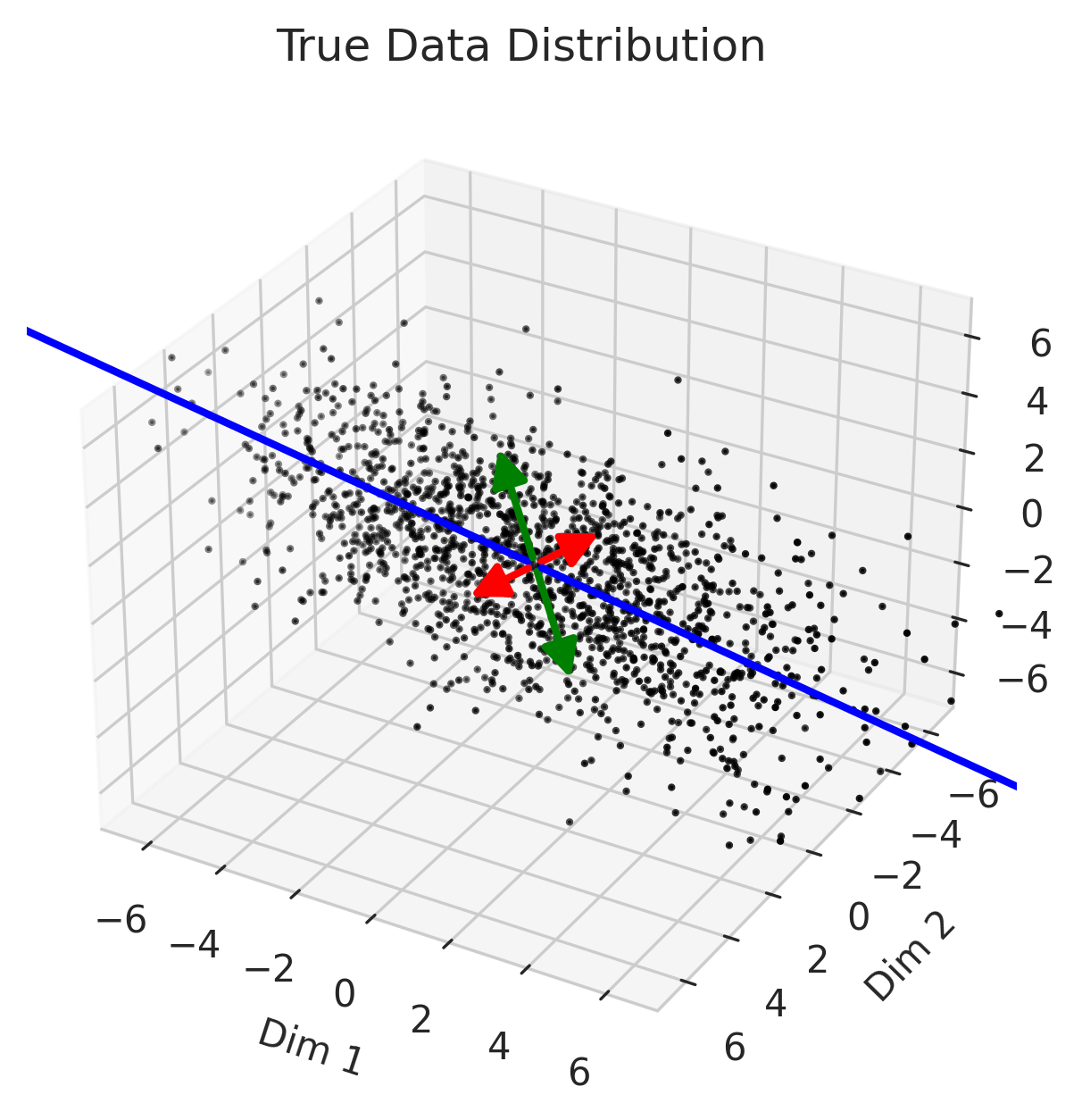}%
    \includegraphics[width=0.32\textwidth]{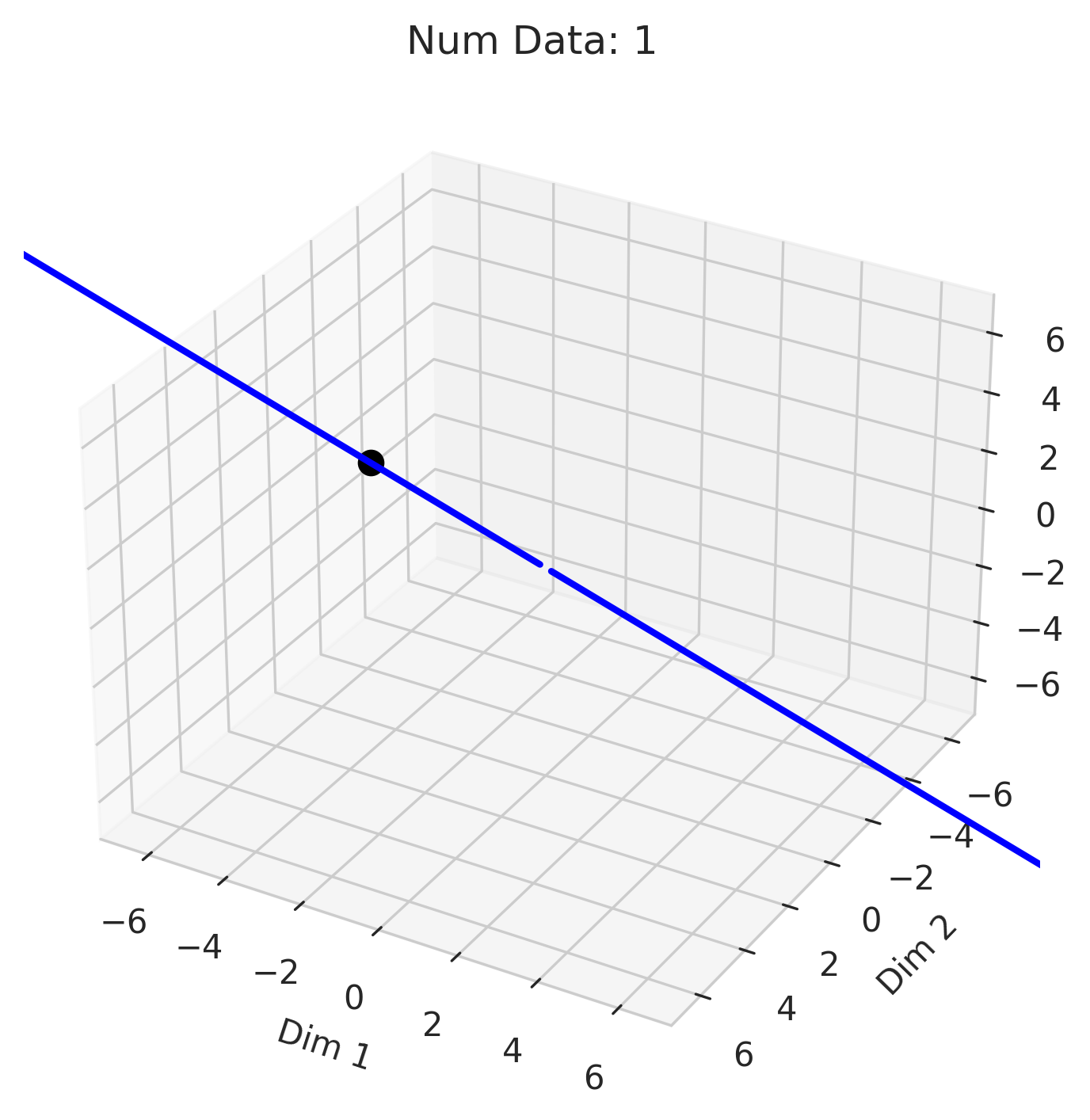}%
    \includegraphics[width=0.32\textwidth]{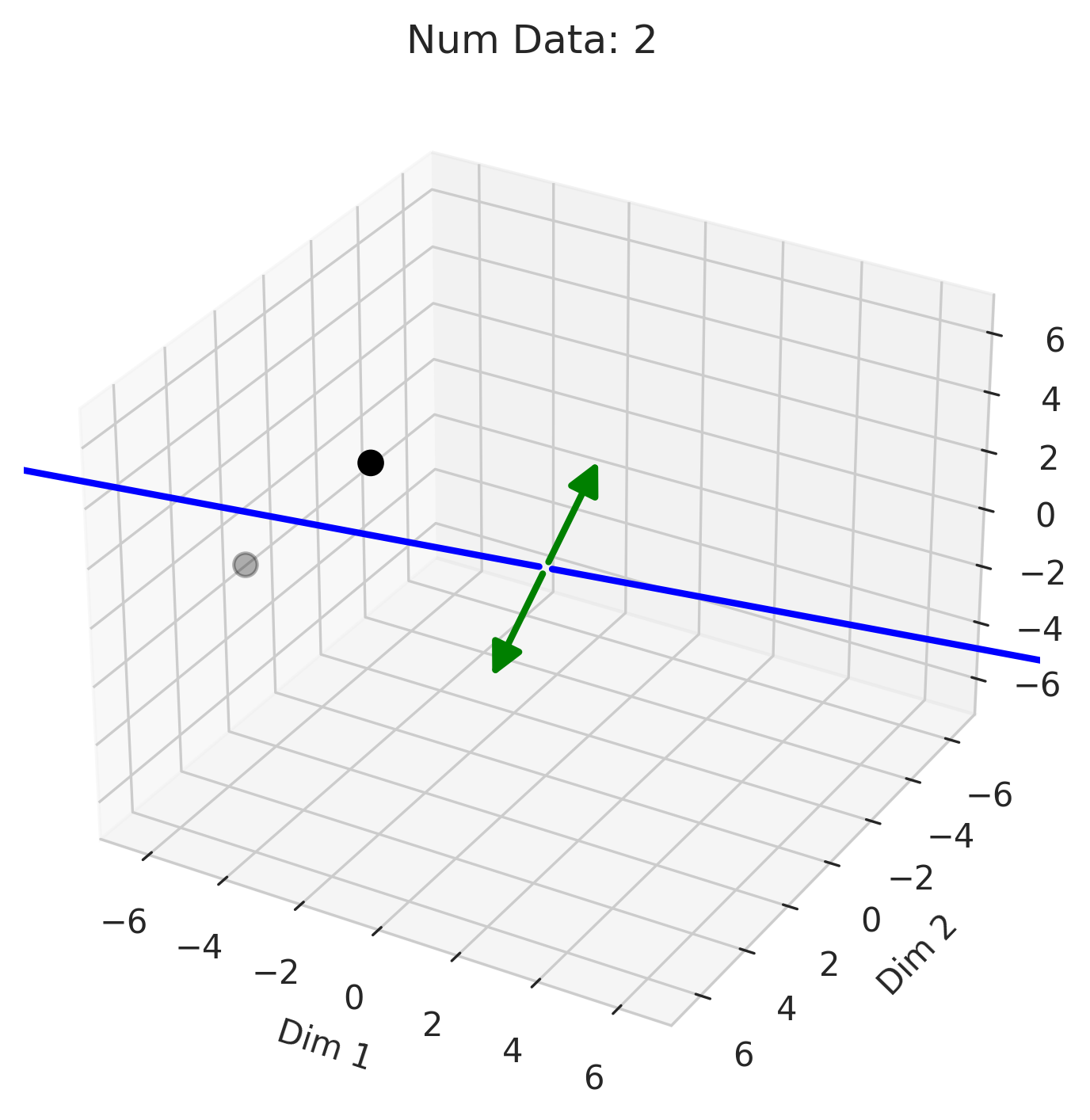}
    \includegraphics[width=0.32\textwidth]{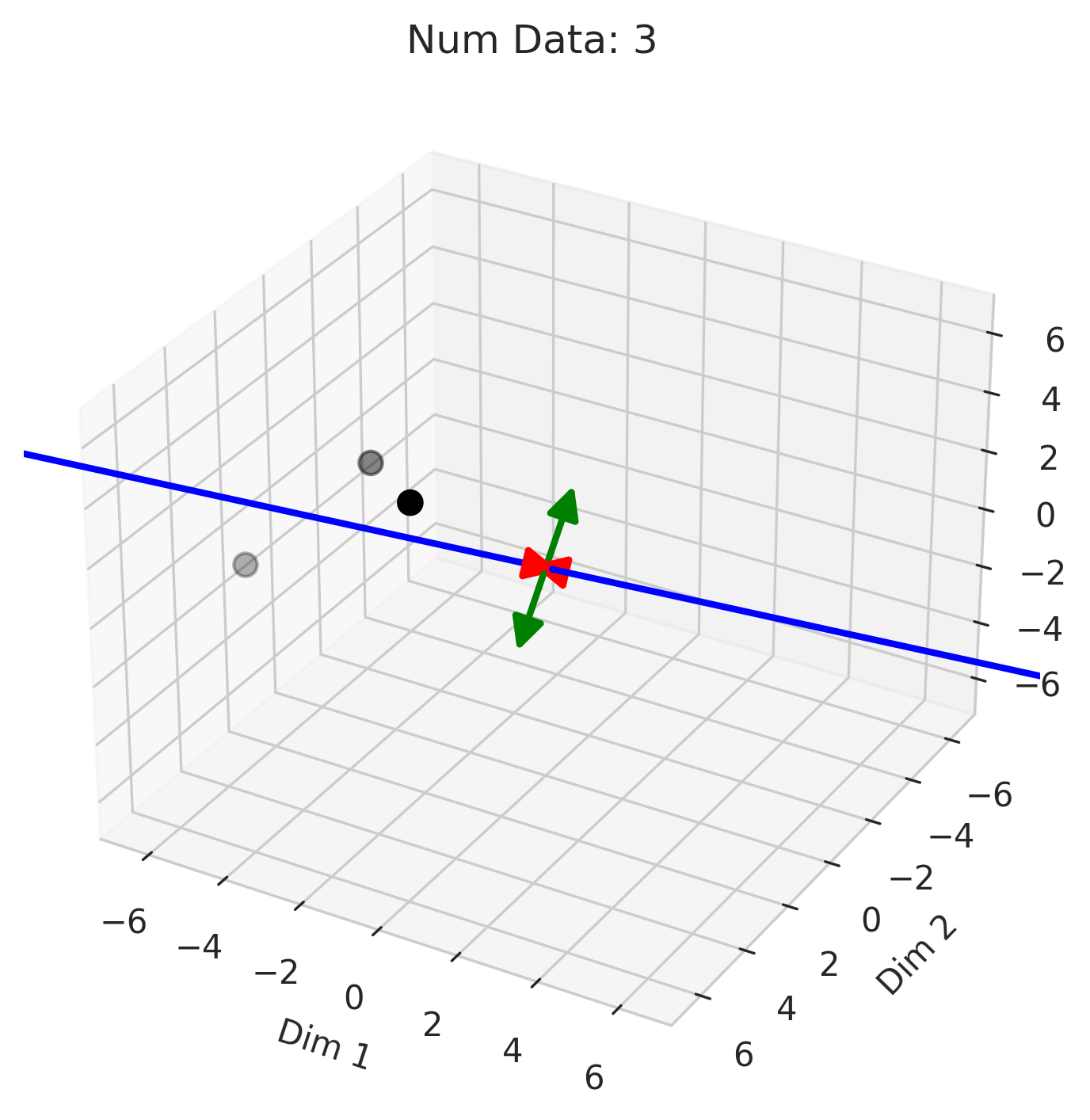}%
    \includegraphics[width=0.32\textwidth]{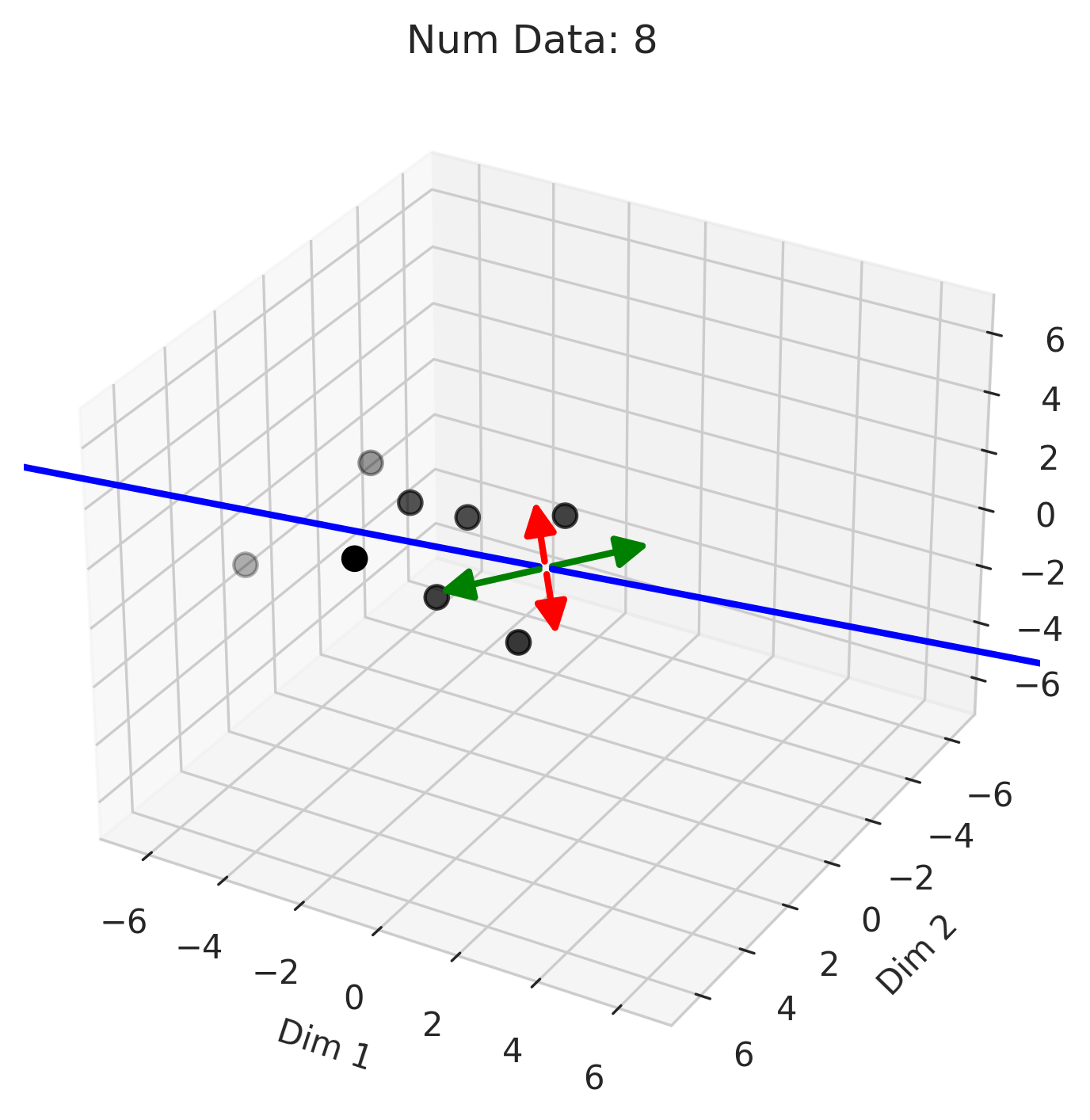}%
    \includegraphics[width=0.32\textwidth]{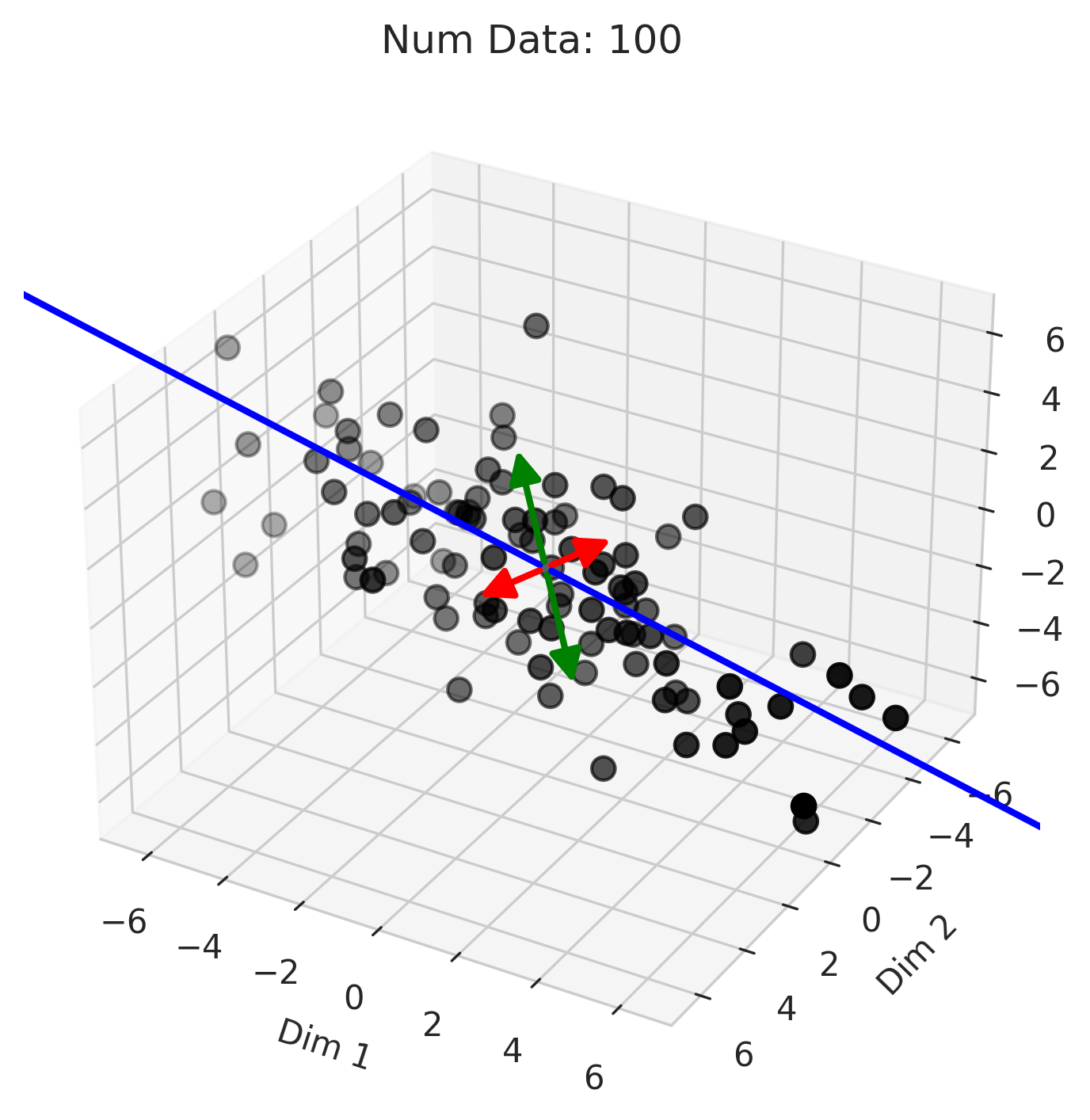}%
    \caption{\textbf{Geometric intuition for why the smallest non-zero singular value reaches its lowest value when approaching the interpolation threshold.} As one nears the interpolation threshold (here, $D=P=N=3$), $N$ training data are unlikely to vary substantially in $P$ orthogonal directions, meaning at least one orthogonal direction is likely to have small variance. If fewer data are observed, then the trailing directions have exactly zero variance, and if more data are observed, then the additional data reveal variance along these trailing directions.}
    \label{fig:geometric_viewpoint}
\end{figure}

Why does this explosion happen near the interpolation threshold? The answer is that the first factor becomes more likely to occur when approaching the interpolation threshold from either parameterization regime. The reason why the smallest non-zero singular value is (probabilistically) likely to reach its lowest value at the interpolation threshold is a probabilistic one, based on the \href{https://en.wikipedia.org/wiki/Marchenko\%E2\%80\%93Pastur\_distribution}{Marchenko–Pastur distribution} from random matrix theory. Because the Marchenko–Pastur distribution is rather technical, we instead focus on gaining intuition by thinking about how much variance we've seen along each orthogonal direction in the data feature space (Fig. \ref{fig:geometric_viewpoint}).\newline

\begin{figure}
    \centering
    \includegraphics[width=0.48\textwidth]{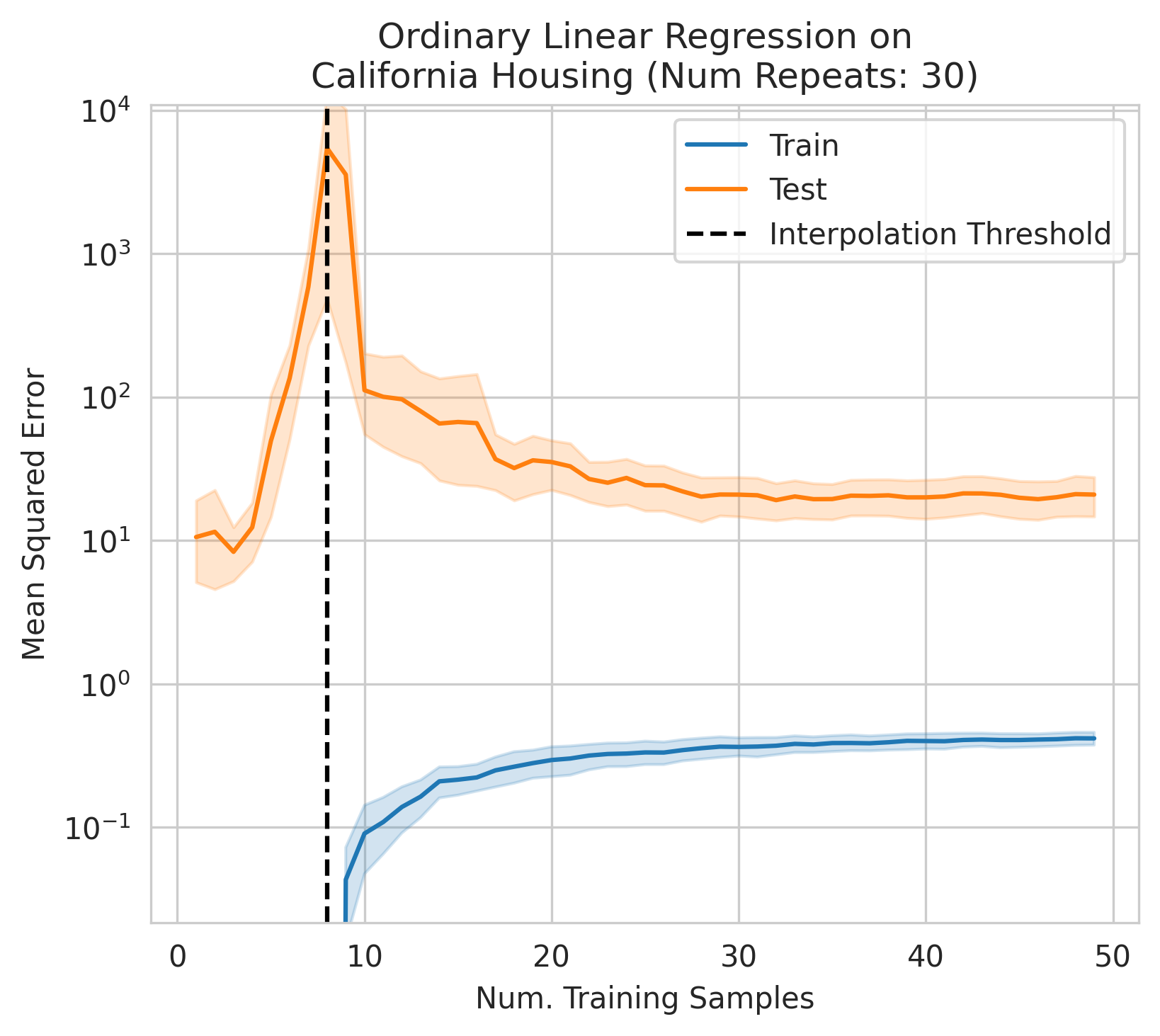}%
    \includegraphics[width=0.49\textwidth]{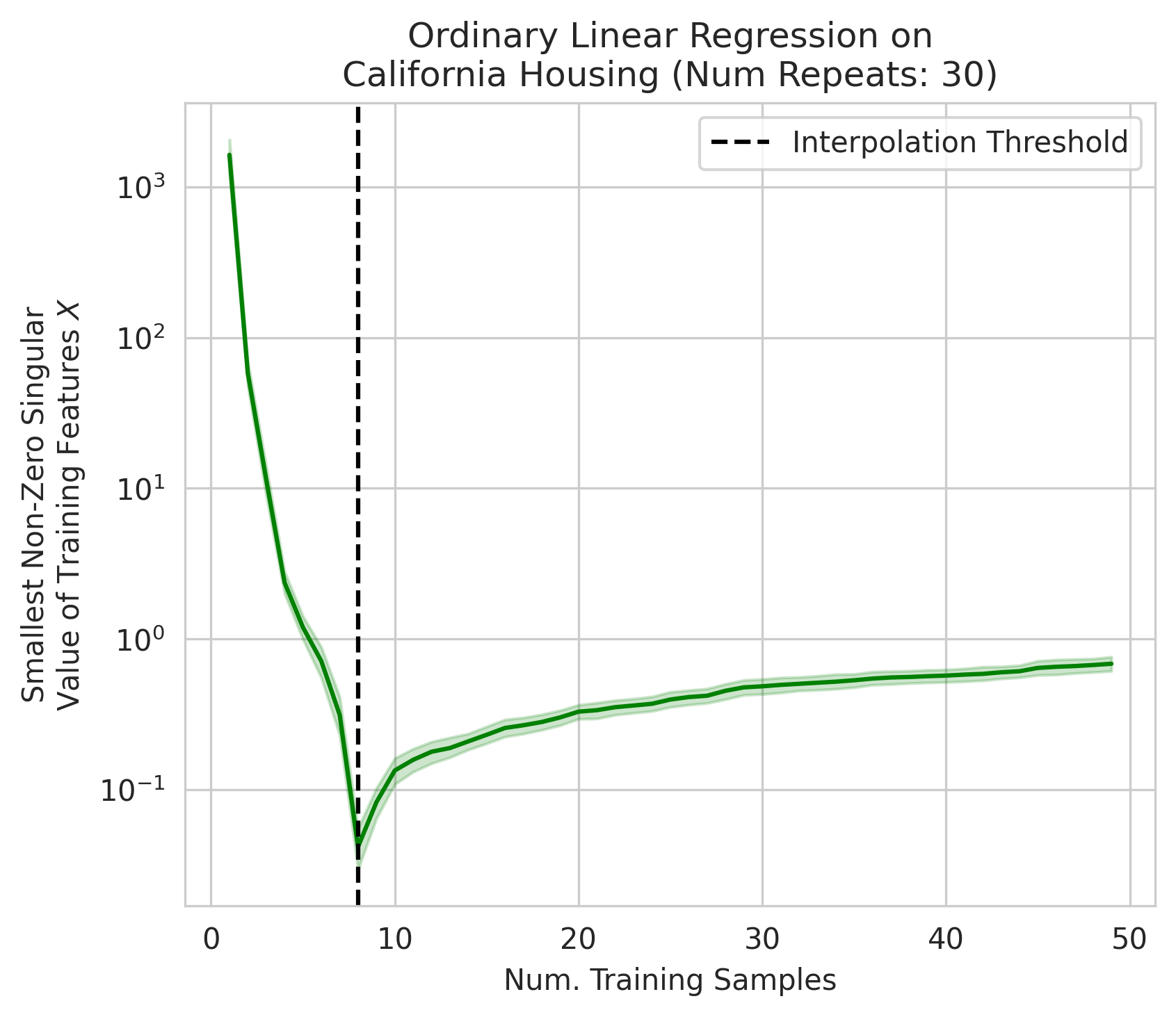}
    \caption{Double Descent in Ordinary Linear Regression on California Housing Dataset.}
    \label{fig:California_housing}
\end{figure}

Suppose we're given a single training datum $\vec{x}_1$. So long as this datum isn't exactly zero, that datum varies in a single direction, meaning we gain information about the data distribution's variance in that direction. Of course, the variance in all orthogonal directions is exactly 0, which the linear regression fit will ignore. Now, suppose we're given a second training datum $\vec{x}_2$. Again, so long as this datum isn't exactly zero, that datum varies, but now, some fraction of $\vec{x}_2$ might have a positive projection along $\vec{x}_1$; if this happens (and it likely will, since the two vectors are unlikely to be exactly orthogonal), the shared direction of the two vectors gives us \textit{more} information about the variance in this shared direction, but gives us \textit{less} information about the second orthogonal direction of variation. This means that the training data's smallest non-zero singular value after 2 samples is probabilistically smaller than after 1 sample. As we gain more training data, thereby approaching the interpolation threshold, the probability that each additional datum has large variance in a new direction orthogonal to all previously seen directions grows increasingly unlikely. At the interpolation threshold, where $N = P = D$, in order for the $N$-th datum to avoid adding a small-but-nonzero singular value to the training data, two properties must hold: (1) there must be one dimension that none of the preceding $N-1$ training data varied in, and (2) the $N$-th datum needs to vary significantly in this single dimension. That's pretty unlikely! As we move beyond the interpolation threshold, the variance in each covariate dimension becomes increasingly clear, and the smallest non-zero singular values moves away from 0. This is displayed visually in Fig. \ref{fig:geometric_viewpoint}.



\begin{figure}
    \centering
    \includegraphics[width=0.48\textwidth]{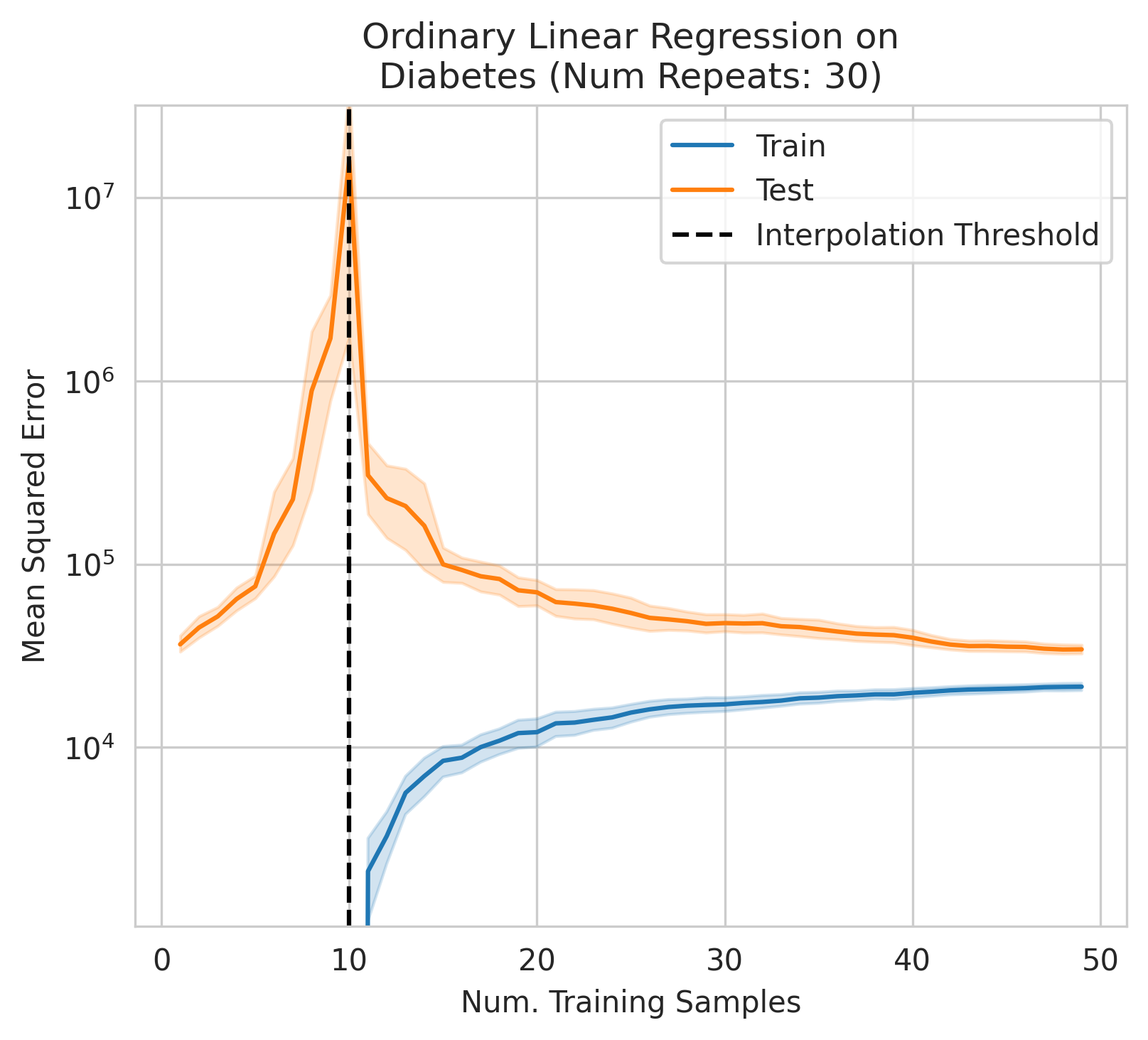}
    \includegraphics[width=0.49\textwidth]{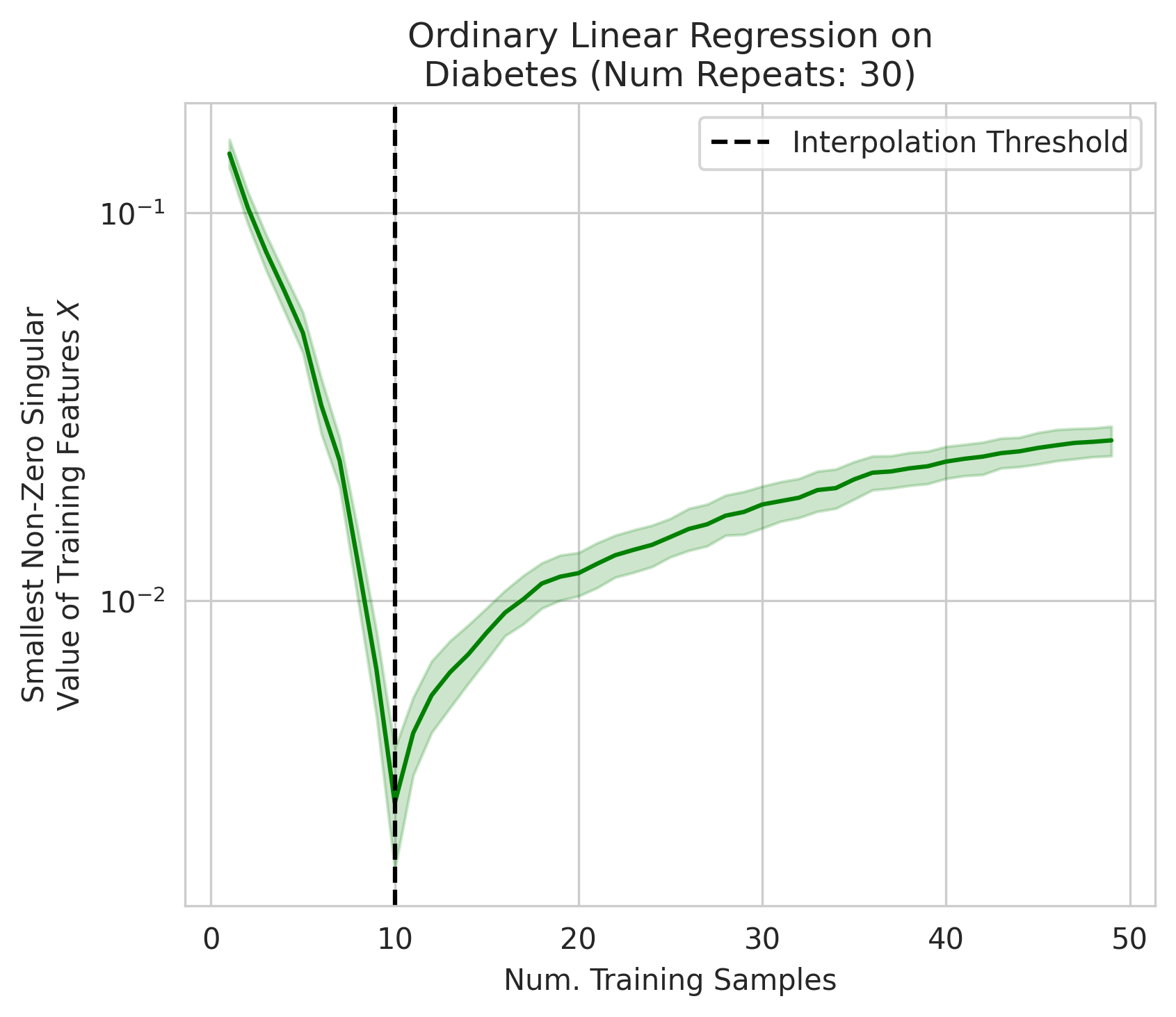}
    \caption{Double Descent in Ordinary Linear Regression on Diabetes Dataset.}
    \label{fig:Diabetes}
\end{figure}

\section{Empirical Evidence on Real Data with Linear Regression}

Does our claim that ordinary linear regression can exhibit double descent hold empirically? We show that it indeed does, using three datasets: WHO Life Expectancy, California Housing, Diabetes. These three datasets were randomly selected on the basis of being easily accessible, e.g., through sklearn \cite{scikit-learn}. All display a sharp spike in test mean squared error at the interpolation threshold (Left panels of Figs. \ref{fig:WHO_life_expectancy}, \ref{fig:California_housing}, \ref{fig:Diabetes}). We additionally substantiate our previous claim that the smallest non-zero singular value of the training features $X$ obtains its lowest value as one nears the interpolation threshold (Right panels of Figs. \ref{fig:WHO_life_expectancy}, \ref{fig:California_housing}, \ref{fig:Diabetes}).
Code is \href{https://github.com/RylanSchaeffer/Stanford-AI-Alignment-Double-Descent-Tutorial}{publicly available}.

\section{When Does Double Descent Not Occur?}
\label{sec:double_descent_ablation}

Double descent will not occur if any of the three factors are absent. What could cause that?

\begin{itemize}
    \item \textit{Small-but-nonzero singular values do not appear in the training data features}. One way to accomplish this is by switching from ordinary linear regression to ridge regression, which effectively adds a gap separating the smallest non-zero singular value from $0$.
    \item \textit{The test datum does not vary in different directions than the training features}. If the test datum lies entirely in the subspace of just a few of the leading singular directions, then double descent is unlikely to occur.
    \item \textit{The best possible model in the model class makes no errors on the training data.} For instance, suppose we use a linear model class on data where the true relationship is a noiseless linear one. Then, at the interpolation threshold, we will have $D=P$ data, $P=D$ parameters, our line of best fit will exactly match the true relationship, and no double descent will occur.
\end{itemize}

To confirm our understanding, we causally test the predictions of when double descent will not occur by ablating each of the three factors individually. Specifically, we do the following:

\begin{enumerate}
    \item No Small Singular Values in Training Features: As we run the ordinary linear regression fitting process, as we sweep the number of training data, we also sweep different singular value cutoffs and remove all singular values of the training features $X$ below the cutoff.
    \item Test Features Lie in the Training Features Subspace: As we run the ordinary linear regression fitting process, as we sweep the number of training data, we project the test features $\vec{x}_{test}$ onto the subspace spanned by the training features $X$ singular modes.
    \item No Residual Errors in the Optimal Model: We first use the entire dataset to fit a linear model $\vec{\beta}^*$, then replace $Y$ with $X \vec{\beta}^*$ and $y_{test}^*$ with $\vec{x}_{test} \cdot \vec{\beta}^*$ to ensure the true relationship is linear. We then rerun our typical fitting process, sweeping the number of training data.
\end{enumerate}

\begin{figure}
    \centering
    \includegraphics[width=0.99\textwidth]{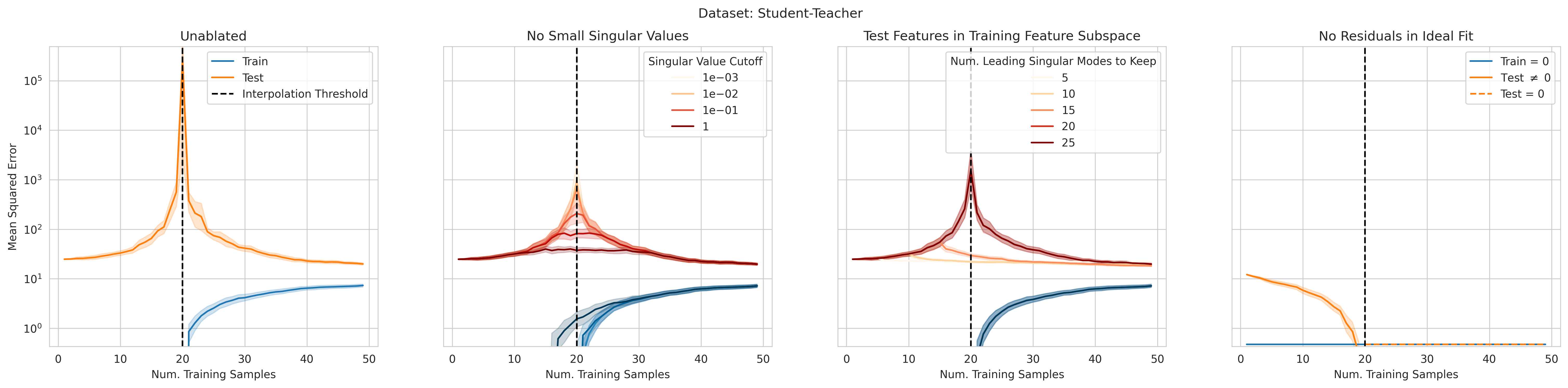}
    \includegraphics[width=0.99\textwidth]{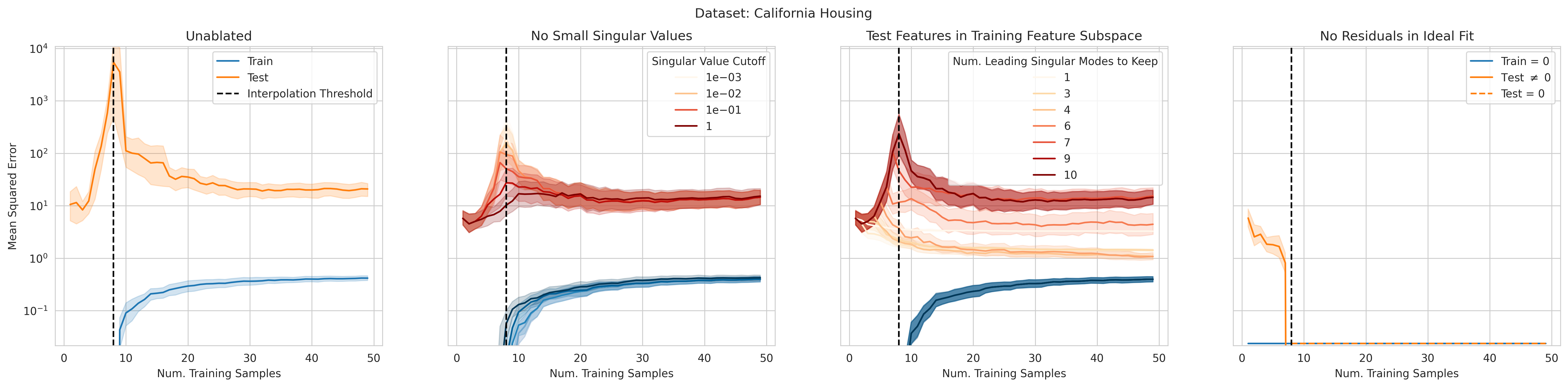}
    \includegraphics[width=0.99\textwidth]{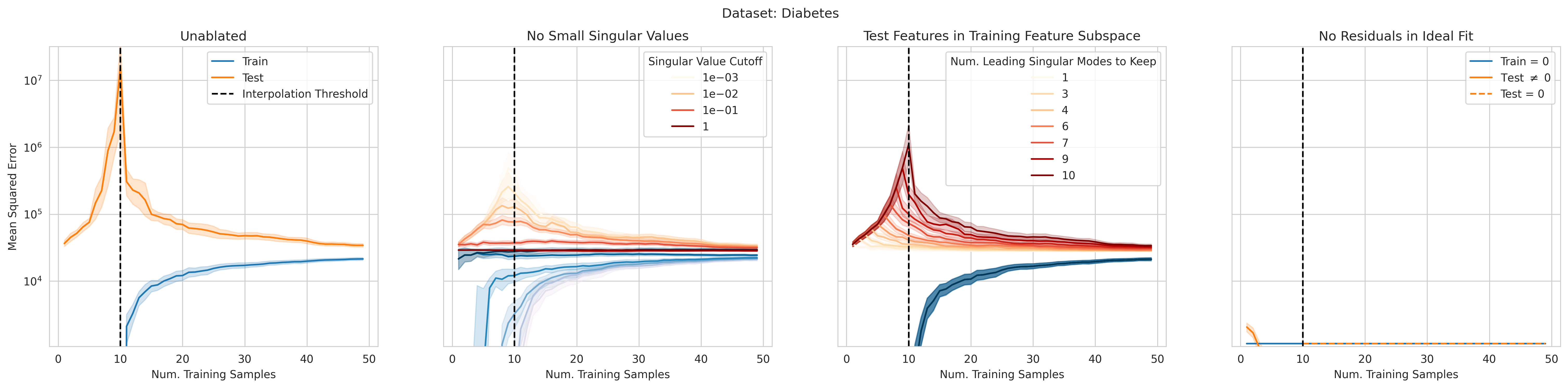}
    \includegraphics[width=0.99\textwidth]{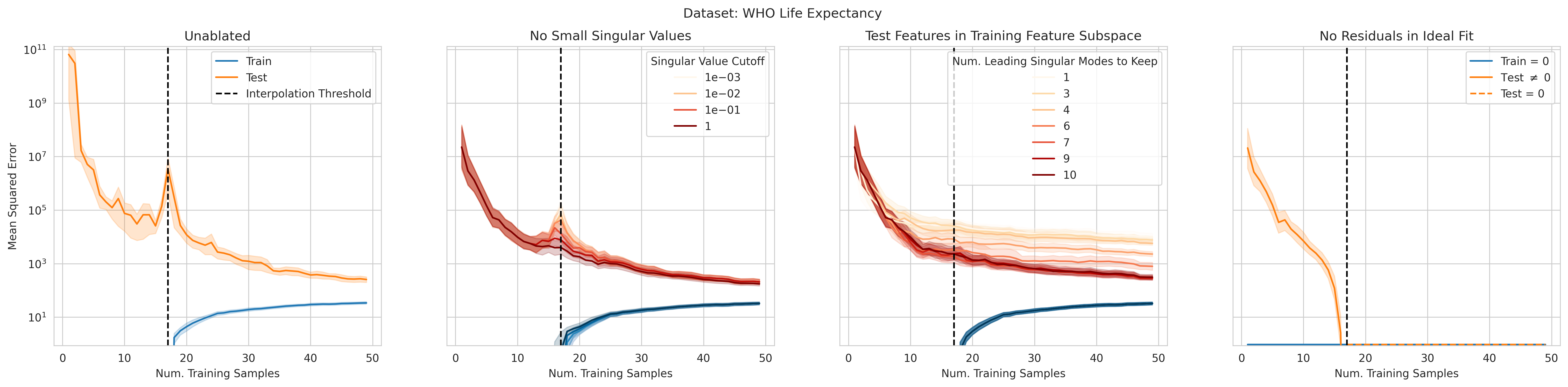}
    \caption{\textbf{Double descent will not occur if any of the three critical quantities are absent.} We demonstrate this via ablations (Sec. \ref{sec:double_descent_ablation}). Left to Right: Double descent appears in ordinary linear regression. Removing small singular values in the training features $X$ prevents double descent. Preventing the test features $\vec{x}_{test}$ from varying in the trailing singular modes of the training features $X$ prevents double descent. Ensuring that the optimal model in the model class has zero residual prediction errors $E$ prevents double descent. Top to Bottom: Synthetic data generated in a Student-Teacher framework, California Housing dataset, Diabetes dataset, WHO Life Expectancy dataset}
    \label{fig:double_descent_ablation}
\end{figure}

We first conduct experiments on a synthetic dataset in a student-teacher setup, and find that causally ablating each of the three factors prevents double descent from occurring (Fig. \ref{fig:double_descent_ablation}, top row). Next, we apply the same ablations to real world datasets (California Housing, Diabetes, WHO Life Expectancy) and find in all three that removing any of the three factors prevents double descent (Fig. \ref{fig:double_descent_ablation}, rows 2-5).

%% file: 04_nonlinear.tex
\section{Intuition Extends to Nonlinear Models}

Although we mathematically studied ordinary linear regression, the intuition for why double descent occurs extends to nonlinear models, including deep neural networks.
For instance, we can conduct the same analysis we did for linear regression but instead for polynomial regression by considering the training covariates $X$ projected into a (typically much higher dimensional) polynomial feature space: 
\begin{equation*}
    \Phi_P(X) \, \defeq \, \begin{bmatrix}
    -\Vec{\phi}_P(\vec{x}_1)-\\
    -\Vec{\phi}_P(\vec{x}_2)-\\
    \vdots\\
    -\Vec{\phi}_P(\vec{x}_N)-\\
    \end{bmatrix}
\end{equation*}

For certain classes of infinitely wide deep neural networks (e.g., as studied \cite{jacot2018neural, lee2017deep, bordelon2020spectrum}), the output predictions of the network are equivalent to a linear regression performed on a particular set of features. Different infinite width limits correspond to different sets of features; for instance, one such limit (known as the Neural Network Gaussian Process limit \cite{lee2017deep}), studies the training covariates $X$ represented in the penultimate layer of the deep neural network $\vec{f}_{\theta}(x)$:
\begin{equation*}
    F_{\theta}(X) \, \defeq \, \begin{bmatrix}
    -\Vec{f}_{\theta}(\vec{x}_1)-\\
    -\Vec{f}_{\theta}(\vec{x}_2)-\\
    \vdots\\
    -\Vec{f}_{\theta}(\vec{x}_N)-\\
    \end{bmatrix}
\end{equation*}

For a concrete example about how our intuition can shed light on the behavior of nonlinear models, \cite{henighan2023superposition} recently discovered interesting properties of shallow nonlinear autoencoders: depending on the number of training data, (1) autoencoders either store data points or features, and (2) double descent occurs between these two regimes. Our tutorial helps explains the results, and also sheds light on two comments the authors make:

\begin{enumerate}
    \item \cite{henighan2023superposition} write, ``[Our work] suggests a naive mechanistic theory of overfitting and memorization: memorization and overfitting occur when models operate on ``data point features" instead of "generalizing features". We expect this naive theory to be overly simplistic, but it seems possible that it's gesturing at useful principles!" Our tutorial hopefully clarifies that this choice of terminology (``data point features" vs. "generalizing features") can be made more precise. When overparameterized, the ``data point features" are akin to the data-by-data Gram matrix $X X^T \in \mathbb{R}^{N \times N}$ and when underparameterized, the ``generalizing features" are akin to the feature-by-feature second moment matrix $X^T X \in \mathbb{R}^{D \times D}$. Our tutorial hopefully shows that ``data point features" can (and very often do) generalize, and that there is a deep connection between the two, e.g., their shared spectra. 
    \item \cite{henighan2023superposition} write, ``It’s interesting to note that we’re observing double-descent in the absence of label noise. That is to say: the inputs and targets are exactly the same. Here, the “noise” arises from the lossy compression happening in the down projection." Our tutorial clarifies that noise in the sense of a random unpredictable quantity is \textit{not} necessary to produce double descent. Rather, what is necessary is \textit{residual errors from the perspective of the model class}. Those residual errors could be entirely deterministic, such as a nonlinear model attempting to fit a noiseless linear relationship.
\end{enumerate}

%% file: 05_discussion.tex
\section{Acknowledgements}

We thank David K. Zhang, Matthew Sotoudeh, Victor Lecomte, Krista Opsahl-Ong, Aaron Scher, Max Lamparth, Zane Durante, Gabriel Mukobi, Brando Miranda and Laureline Logiaco for feedback.

%% file: Implicit_Bias_GD.tex


\section{Why Gradient Descent Implicitly Regularizes}
\label{app:why_sgd_regularizes}

This is a sketch of why gradient descent implicitly regularizes. Suppose we have a model $X w$ for a vector of data $y \in \mathbb{R}^n$ and want to minimize the norm of the error,
$$
L(w) = \Vert X w - y \Vert_2^2 = \Vert e \Vert_2^2
$$
where we introduce some short-hand notation. We use the gradient learning rule,
$$
w(t+1) = w(t) - \eta X^T e(t) $$
$$
\Rightarrow e(t+1) = e(t) - \eta X X^T e(t)  $$
$$
\Rightarrow e(t+1) = (I - \eta X X^T) e(t)
$$
Each matrix satisfies $X \in \mathbb{R}^{n \times d_1}$ where $n$ is the number of samples and $d_1$ is the dimension of each sample. In the overparameterized setting we have $d_1 > n$ and so $X X^T$ will generically have full-rank and the error will go to zero. 

This lies in the difference between $X X^T$ which appears here in the error analysis and $X^T X$ which appears in the solution. So we can have $X X^T \in \mathbb{R}^{n \times n}$ generically full-rank only if we have more parameters than there is data. On the other hand, we only have $X^T X$ full-rank if also it's satisfied that there is more data than parameters. This is important because in this case we can compute the pseudo-inverse easily. Generically, we can show that if we use gradient descent we have something like the following,
$$
\underbrace{(X^T X)^{-1} X}_{\text{left inverse}} \quad \underbrace{X^{-1}}_{\text{inverse}} \quad \underbrace{X^T (X X^T)^{-1}}_{\text{right inverse}}
$$
for the cases where we are under-parameterized, minimally parameterized, or over-parameterized to model the data. 

So under gradient flow we'll suppose the parameters update according to,
$$
\dot w = - \eta X^T e$$
$$w(0) = 0
$$
Observe that the gradient $\dot w$ is invariantly in the span of $X^T$ so we may conclude that $w(t)$ is always in the span of $X^T$. Generically, any solution in the over-parameterized setting is a global optimizer such that $X w = y$. This means that the limit of the flow can be written as $w^* = X^T \alpha$ for some coefficient vector with the constraint that $X w^* = y$. After some manipulations we find that,
$$
y = X w^* = X X^T \alpha $$ $$
\Rightarrow \alpha = (X X^T)^{-1} y $$ $$
\Rightarrow w^* = X^T (X X^T)^{-1} y = X^+ y
$$
This means that the solution $X^+$ picked from gradient flow is the Moore-Penrose psuedoinverse. This can be defined as the matrix,
$$
X^+ = \lim_{\lambda \to 0^+} X^T (X X^T + \lambda  I)^{-1}
$$
Also observe that there is a unique minimizer for the regularized problem,
$$
\min_w L(w) + \lambda \Vert w \Vert_2^2
$$
with value $w_{\lambda} = X^T (X X^T + \lambda I)^{-1} y$. Perhaps, $X w = y$ has a set of solutions, but it is clear this set is convex so there is a unique minimum norm solution. On the other hand, each $w_{\lambda}$ corresponds to a best solution with norm below $v_{\lambda}$, which is less than the minimum. However, we have $w^* = \lim_{\lambda \to 0^+} w_{\lambda}$ from continuity. Since $w^*$ is an exact solution it can't have less than the minimum-norm and it is clear $w^*$ can't have above the minimum-norm either since this is not the case for any of the $w_{\lambda}$. We conclude that gradient descent does indeed find the minimum norm solution.

%% file: 00_main.bbl
\begin{thebibliography}{10}

\bibitem{adlam2020understanding}
Ben Adlam and Jeffrey Pennington.
\newblock Understanding double descent requires a fine-grained bias-variance
  decomposition.
\newblock {\em Advances in neural information processing systems},
  33:11022--11032, 2020.

\bibitem{advani2020high}
Madhu~S Advani, Andrew~M Saxe, and Haim Sompolinsky.
\newblock High-dimensional dynamics of generalization error in neural networks.
\newblock {\em Neural Networks}, 132:428--446, 2020.

\bibitem{belkin2019reconciling}
Mikhail Belkin, Daniel Hsu, Siyuan Ma, and Soumik Mandal.
\newblock Reconciling modern machine-learning practice and the classical
  bias--variance trade-off.
\newblock {\em Proceedings of the National Academy of Sciences},
  116(32):15849--15854, 2019.

\bibitem{bordelon2020spectrum}
Blake Bordelon, Abdulkadir Canatar, and Cengiz Pehlevan.
\newblock Spectrum dependent learning curves in kernel regression and wide
  neural networks.
\newblock In {\em International Conference on Machine Learning}, pages
  1024--1034. PMLR, 2020.

\bibitem{chen2021multiple}
Lin Chen, Yifei Min, Mikhail Belkin, and Amin Karbasi.
\newblock Multiple descent: Design your own generalization curve.
\newblock {\em Advances in Neural Information Processing Systems},
  34:8898--8912, 2021.

\bibitem{hastie2022surprises}
Trevor Hastie, Andrea Montanari, Saharon Rosset, and Ryan~J Tibshirani.
\newblock Surprises in high-dimensional ridgeless least squares interpolation.
\newblock {\em The Annals of Statistics}, 50(2):949--986, 2022.

\bibitem{henighan2023superposition}
Tom Henighan, Shan Carter, Tristan Hume, Nelson Elhage, Robert Lasenby,
  Stanislav Fort, Nicholas Schiefer, and Christopher Olah.
\newblock Double descent in the condition number.
\newblock {\em Transformer Circuits Thread}, 2023.

\bibitem{jacot2018neural}
Arthur Jacot, Franck Gabriel, and Cl{\'e}ment Hongler.
\newblock Neural tangent kernel: Convergence and generalization in neural
  networks.
\newblock {\em Advances in neural information processing systems}, 31, 2018.

\bibitem{lee2017deep}
Jaehoon Lee, Yasaman Bahri, Roman Novak, Samuel~S Schoenholz, Jeffrey
  Pennington, and Jascha Sohl-Dickstein.
\newblock Deep neural networks as gaussian processes.
\newblock {\em arXiv preprint arXiv:1711.00165}, 2017.

\bibitem{mei2022generalization}
Song Mei and Andrea Montanari.
\newblock The generalization error of random features regression: Precise
  asymptotics and the double descent curve.
\newblock {\em Communications on Pure and Applied Mathematics}, 75(4):667--766,
  2022.

\bibitem{nakkiran2021deep}
Preetum Nakkiran, Gal Kaplun, Yamini Bansal, Tristan Yang, Boaz Barak, and Ilya
  Sutskever.
\newblock Deep double descent: Where bigger models and more data hurt.
\newblock {\em Journal of Statistical Mechanics: Theory and Experiment},
  2021(12):124003, 2021.

\bibitem{nakkiran2020optimal}
Preetum Nakkiran, Prayaag Venkat, Sham Kakade, and Tengyu Ma.
\newblock Optimal regularization can mitigate double descent.
\newblock {\em arXiv preprint arXiv:2003.01897}, 2020.

\bibitem{opper1995statistical}
Manfred Opper.
\newblock Statistical mechanics of learning: Generalization.
\newblock {\em The handbook of brain theory and neural networks}, pages
  922--925, 1995.

\bibitem{scikit-learn}
F.~Pedregosa, G.~Varoquaux, A.~Gramfort, V.~Michel, B.~Thirion, O.~Grisel,
  M.~Blondel, P.~Prettenhofer, R.~Weiss, V.~Dubourg, J.~Vanderplas, A.~Passos,
  D.~Cournapeau, M.~Brucher, M.~Perrot, and E.~Duchesnay.
\newblock Scikit-learn: Machine learning in {P}ython.
\newblock {\em Journal of Machine Learning Research}, 12:2825--2830, 2011.

\bibitem{poggio2019double}
Tomaso Poggio, Gil Kur, and Andrzej Banburski.
\newblock Double descent in the condition number.
\newblock {\em arXiv preprint arXiv:1912.06190}, 2019.

\bibitem{rocks2021geometry}
Jason~W Rocks and Pankaj Mehta.
\newblock The geometry of over-parameterized regression and adversarial
  perturbations.
\newblock {\em arXiv preprint arXiv:2103.14108}, 2021.

\bibitem{rocks2022bias}
Jason~W Rocks and Pankaj Mehta.
\newblock Bias-variance decomposition of overparameterized regression with
  random linear features.
\newblock {\em Physical Review E}, 106(2):025304, 2022.

\bibitem{rocks2022memorizing}
Jason~W Rocks and Pankaj Mehta.
\newblock Memorizing without overfitting: Bias, variance, and interpolation in
  overparameterized models.
\newblock {\em Physical Review Research}, 4(1):013201, 2022.

\bibitem{spigler2018jamming}
Stefano Spigler, Mario Geiger, St{\'e}phane d'Ascoli, Levent Sagun, Giulio
  Biroli, and Matthieu Wyart.
\newblock A jamming transition from under-to over-parametrization affects loss
  landscape and generalization.
\newblock {\em arXiv preprint arXiv:1810.09665}, 2018.

\end{thebibliography}
